\documentclass[lettersize,journal]{IEEEtran}
\usepackage{svg}
\usepackage{cite}
\usepackage{amsmath,amssymb,amsfonts}
\usepackage{algorithmic}
\usepackage{graphicx} 
\usepackage{subcaption}
\usepackage{textcomp}
\usepackage{xcolor}
\usepackage[linesnumbered,ruled,vlined]{algorithm2e}
\usepackage{xcolor}  
\usepackage{subcaption}
\usepackage{dblfloatfix}
\usepackage{color} 
\usepackage{accents} 
\usepackage{multirow}
\usepackage{tabularray}
\newcommand{\bfmath}[1]{\mbox{\boldmath$#1$\unboldmath}}
\def\BibTeX{{\rm B\kern-.05em{\sc i\kern-.025em b}\kern-.08em
    T\kern-.1667em\lower.7ex\hbox{E}\kern-.125emX}}
 
\begin{document}

\title{An Efficient Continual Learning Framework for Multivariate Time Series Prediction Tasks with Application to Vehicle State Estimation }

\author{Arvin Hosseinzadeh, Ladan Khoshnevisan, Mohammad Pirani, \\ Shojaeddin Chenouri, Amir Khajepour ~\IEEEmembership{}
\thanks{Arvin Hosseinzadeh, Ladan Khoshnevisan, and Amir Khajepour are with the Department of Mechanical and Mechatronics Engineering, University of Waterloo, Waterloo, ON N2L 3G1, Canada (e-mail: arvin.hosseinzadeh@uwaterloo.ca; lkhoshnevisan@uwaterloo.ca; a.khajepour@uwaterloo.ca)} 
\thanks{Mohammad Pirani is with the Department of Mechanical Engineering, University of Ottawa, Ottawa, Ontario ON K1N 6N5, Canada (e-mail: mpirani@uottawa.ca)} 
\thanks{Shojaeddin Chenouri is with the Department of Statistics and Actuarial Science, University of Waterloo, Waterloo, ON N2L 3G1, Canada (e-mail: schenouri@uwaterloo.ca)}

\thanks{\textit{Corresponding author: Arvin Hosseinzadeh}}
}

\markboth{}%
{Shell \MakeLowercase{\textit{et al.}}: A Sample Article Using IEEEtran.cls for IEEE Journals}

\IEEEpubid{}

\maketitle

\begin{abstract}
In continual time series analysis using neural networks, catastrophic forgetting (CF) of previously learned models when training on new data domains has always been a significant challenge. This problem is especially challenging in vehicle estimation and control, where new information is sequentially introduced to the model. Unfortunately, existing work on continual learning has not sufficiently addressed the adverse effects of catastrophic forgetting in time series analysis, particularly in multivariate output environments. In this paper, we present EM-ReSeleCT (Efficient Multivariate Representative Selection for Continual Learning in Time Series Tasks), an enhanced approach designed to handle continual learning in multivariate environments. Our approach strategically selects representative subsets from old and historical data. It incorporates memory-based continual learning techniques with an improved optimization algorithm to adapt the pre-trained model on new information while preserving previously acquired information. Additionally, we develop a sequence-to-sequence transformer model (autoregressive model) specifically designed for vehicle state estimation. Moreover, we propose an uncertainty quantification framework using conformal prediction to assess the sensitivity of the memory size and to showcase the robustness of the proposed method. Experimental results from tests on an electric Equinox vehicle highlight the superiority of our method in continually learning new information while retaining prior knowledge, outperforming state-of-the-art continual learning methods. Furthermore, EM-ReSeleCT significantly reduces training time, a critical advantage in continual learning applications.    
\end{abstract}


\section{Introduction}
\IEEEPARstart{M}{achine} learning models have emerged as powerful tools for addressing complex problems in time series predictions \cite{farrell2021deep}. Recently, Neural networks, due to their non-linear modeling capabilities, have become a popular choice for time series analysis and state estimation tasks \cite{hewamalage2021recurrent,jin2024survey}. In \cite{chemali2018state}, a novel approach using deep neural networks was proposed to estimate the state of charge (SOC) of electric vehicles (EV). In \cite{revach2022kalmannet}, a hybrid approach was proposed to use neural networks to aid model-based approaches in state estimation of dynamical systems with partial information. Recently, significant attention has been paid to using Transformer models for time series forecasting tasks due to their computational efficiency and improved estimation accuracy compared to recurrent neural networks (RNNs) \cite{wu2020deep,zhou2021informer}.      

Unfortunately, these machine learning methods frequently struggle in dynamic environments where the model must continuously adapt to new information. \cite{kirkpatrick2017overcoming}. This issue is particularly significant in time series regression problems, such as robot learning, battery management of electric vehicles (EV), and financial market predictions, where models must adapt to new environments containing new information \cite{su2020incremental,eaty2023digital,ramjattan2024comparative}. The challenge becomes even more critical in vehicle state estimation, where learning-based models must continuously adapt to new data domains encountered by vehicles. Existing work on time series state estimation in dynamic and non-stationary environments has largely focused on local models, which apply local regression fits within each model in each data domain to perform estimations \cite{nguyen2009model,schaal2002scalable,liu2024incremental}. However, local models face difficulties in noisy environments and can lead to an overwhelming number of models when many domains are introduced, particularly in large datasets. When using neural networks as an alternative, although more powerful, they present challenges when used with static model parameters. In such cases, a pre-trained model struggles to retain previously learned information when exposed to new data, leading to catastrophic forgetting (CF). Unfortunately, retraining the model on both the new and old data domains is impractical due to the high computational complexity involved. Continual learning (CL) aims to enable models to learn from new information while preserving previously acquired knowledge. 

Several continual learning approaches have been proposed in the literature, which are generally classified into three groups: regularization-based \cite{kirkpatrick2017overcoming,zenke2017continual}, architectural-based \cite{rusu2016progressive}, and memory-based \cite{lopez2017gradient,chaudhry2019tiny,bang2021rainbow,chaudhry2018efficient} methods. In regularization-based methods, important weights or model parameters from previous tasks are preserved to eliminate catastrophic forgetting. Architectural-based methods involve expanding the model's structure as new tasks are introduced. Memory-based continual learning approaches select a subset of historical data to be retained as memory, which is then combined with newly received data for training on the current task. The selected subset must effectively represent the old data.

Unfortunately, all these methods have primarily been applied to classification tasks and struggle to address the issues related to continual learning in time series regression problems. Recently, ReSeleCT \cite{hosseinzadeh2024reselect} was proposed as a method specifically designed to address catastrophic forgetting in state estimation scenarios. Although promising, ReSeleCT has notable limitations. First, it is restricted to univariate cases, focusing on the continual learning and estimation of a single output or state. Second, ReSeleCT employs the exact structure of the average gradient of episodic memory (A-GEM) algorithm \cite{chaudhry2018efficient}, which is a memory-based CL technique and was initially designed for classification tasks and may not be optimal for regression and prediction problems. Third, the original ReSeleCT implementation relies on recurrent neural networks (RNNs) for state estimation, which may be insufficient for more complex time series problems with substantial data volume and intricate patterns. In such scenarios, more advanced models may be needed, as RNNs can struggle in terms of computational efficiency and representational capability.

In this paper, we introduce our modified scheme on the A-GEM algorithm and our approach for multivariate representative or memory selection to eliminate the adverse effects of catastrophic forgetting in multivariate state estimation. In the experiments, we develop a transformer model-based estimation with an encoder-decoder architecture to improve estimation tasks. Originally introduced in \cite{vaswani2017attention}, transformer models were designed to enhance the performance of large-language models as an alternative to traditional RNN methods. In this work, we adapt transformer models specifically for state estimation tasks, particularly adapted for vehicle datasets. The key contributions in this study are outlined as follows:

\begin{itemize}
    \item A simple yet effective modification to the A-GEM algorithm to enhance the preservation of historical data, which is crucial for continual learning (CL) problems. 

    \item A general framework for continual learning in multivariate time series scenarios. In multivariate time series regression, existing research has predominantly focused on locally based models. Our approach introduces a robust method using state-of-the-art neural network models.
      
    \item We apply experimental analysis on real-world data in the electric Equinox vehicle state estimation to validate the proposed model, adapting a sequence-to-sequence transformer model for vehicle state estimation with an encoder-decoder mechanism.

    \item The integration of conformal prediction to assess and quantify the uncertainty of the proposed method across different memory sizes, with comparisons to the baseline approach. 
\end{itemize}  

The remainder of the paper is organized as follows: Section II presents the problem formulation. Section III introduces A-GEM and our refined approach adapted for time series regression problems. Section IV introduces our approach to continual learning in the multivariate space. Finally, Section V details the experimental analysis of the proposed model in a real-world application of continual learning using a sequence-to-sequence transformer model for vehicle state estimation tasks. Additionally, this section performs an uncertainty quantification of the model using the proposed learning method.

\section{Problem formulation and Background}

We consider a stream of observations denoted as $(\mathbf{x}_i,\,\mathbf{y}_i)$ for $i=1,\,2,\,\dots$, where $\mathbf{y}_i\in\mathbb{R}^q$ represents the desired output or response vector, and $\mathbf{x}_i\in\mathbb{R}^p$ is the input feature vector. The objective is to approximate an unknown $\mathbb{R}^q$ valued function \(f\) based on a dataset $D$ of observations $(\mathbf{x}_i,\,\mathbf{y}_i)$. We assume that for a given integer $d$, which indicates the order of temporal dependency between $(\mathbf{x}_i,\,\mathbf{y}_i)$s and \(i>d\),
\begin{equation}\label{Eq1}
    \mathbf{y}_i = f(\mathbf{x}_i,H_i(D))+\boldsymbol{\epsilon}_i,
\end{equation} 
where \(H_i(D)=\lbrace (\mathbf{x}_j,\,\mathbf{y}_j)\in D;\, i-d\le j<i \rbrace\), and \(\boldsymbol{\epsilon}_i\in\mathbb{R}^q\) is the additive noise. As previously mentioned, neural networks have proven to be highly effective in estimating the function \( f \) for time series problems. By adopting the neural network approach, we inherently assume that \( f \) in \eqref{Eq1} can be adequately estimated by a neural network parameterized by the vector \( \boldsymbol{\theta} \). Therefore, we use \( f(\cdot\,;\,\boldsymbol{\theta}) \) in place of \( f \), effectively reducing the estimation of \( f \) to estimating the parameter vector \( \boldsymbol{\theta} \). Let \( \ell_i(\boldsymbol{\theta}) = \|\mathbf{y}_i - f(\mathbf{x}_i\,;\,\boldsymbol{\theta}))\|^2 \) represent the squared error loss for the \(i\)-th observation, where $\|\cdot\|$ is the $l_2$ norm of the resulting vector. To estimate \( \boldsymbol{\theta} \), it is common to minimize the empirical risk:

\begin{equation}\label{ERisk}
{\rm R}(\boldsymbol{\theta},\,D)=\frac{1}{|\,D\,|}\sum\limits_{i=1}^{|\,D\,|}\ell_i(\boldsymbol{\theta}).
\end{equation} 
where $|\,D\,|$ represents the total number of data points in dataset $D$. Suppose there are $k$ datasets for a given experiment, each comprising time series data structured as in (\ref{Eq1}): 

\begin{equation}\label{AllDataset}
    {\begin{cases}       
D_1 = \left\lbrace \left(\mathbf{x}_1,\,\mathbf{y}_1 \right),\,\dots,\,\left(\mathbf{x}_{n_1},\,\mathbf{y}_{n_1} \right) \right\rbrace\\ 
D_2 = \left\lbrace \left( \mathbf{x}_{n_1+1},\,\mathbf{y}_{n_1+1} \right),\,\dots,\,\left ( \mathbf{x}_{n_2},\,\mathbf{y}_{n_2} \right ) \right \rbrace\\ 
\vdots \\ 
D_k = \left \lbrace \left ( \mathbf{x}_{n_{k-1}+1},\,\mathbf{y}_{n_{k-1}+1} \right ),\,\dots,\,\left ( \mathbf{x}_{n_k},\,\mathbf{y}_{n_k} \right ) \right \rbrace\,
\end{cases}}
\end{equation} 
Here, $n = \sum_{i=1}^k n_i$ represents the total number of data points in the combined dataset. In a continual learning setting, datasets from different domains denoted as $D = D_1 \cup D_2 \cup \dots \cup D_k$, are not available all at once. Instead, new datasets ($D_2, \dots, D_k$) arrive sequentially, requiring the model to update with new data while retaining previously learned information. In memory-based continual learning, after training on an initial dataset \(D_1\), the idea is to select a representative subset of \(m_1\) landmarks, $m_1 \ll n_1$, denoted as \(D^*_{1}\), to retain in memory. Let us consider the selected landmarks from dataset $D_1$ as: 

\begin{equation}\label{AllD_selected}
    D^*_{1} = \left \lbrace \left ( \mathbf{x}^*_{1},\,\mathbf{y}^*_{1} \right ),\,\dots,\,\left (\mathbf{x}^*_{m_1},\,\mathbf{y}^*_{m_1} \right ) \right \rbrace\,.
\end{equation}
This subset is used in conjunction with newly received data \(D_2\) to update the model parameters \(\boldsymbol{\theta}\) by minimizing the empirical risk while ensuring that the risk on the previously selected representatives does not exceed the previous risk:
\begin{align}\label{A-GEM}
\underset{\boldsymbol{\theta}}{\min}\,\,{\rm R}(\boldsymbol{\theta},\,D_2)\quad \text{s.t. }\quad
{\rm R}(\boldsymbol{\theta},\,D_1^\ast)\leq {\rm R}(\widehat{\boldsymbol{\theta}}_1,\,D_1^\ast).
\end{align}

\section{A-GEM and the Refined Approach}
In general, upon receiving the $t$-th dataset $D_t$, one method that minimizes empirical risk while preserving predictive performance on previously learned data is the Average Gradient of Episodic Memory (A-GEM) \cite{chaudhry2018efficient}. This method updates the parameter estimate $\widehat{\bfmath{\theta}}_t$ by solving the following optimization problem:
\begin{align}\label{A-GEM}
\underset{\theta}{\min}\,\,{\rm R}(\bfmath{\theta},\,D_t)\text{ s.t. }
{\rm R}(\bfmath{\theta},\,\bigcup\limits_{i=1}^{t-1}D_i^\ast)\leq {\rm R}(\widehat{\bfmath{\theta}}_{t-1},\,\bigcup\limits_{i=1}^{t-1}D_i^\ast)\,,
\end{align}
resulting in the updated model $\widehat{f}_t = f(\cdot,\,\widehat{\bfmath{\theta}}_t)$. 

In order to apply A-GEM, the gradients are first defined as follows:
\begin{align*}
g^\ast_k&=\frac{\partial}{\partial\theta}{\rm R}(\bfmath{\theta}\,,\,\bigcup_{i=1}^k D^*_{i})\Big|_{\theta=\widehat{\theta}_k}\\
g_{k+1}&=\frac{\partial}{\partial\theta}{\rm R}(\bfmath{\theta}\,,\,D_{k+1})\Big|_{\theta=\widehat{\theta}_k}\,.
\end{align*}

In each epoch of training the model, the procedure includes computing the inner product between the newly obtained gradient $g_{k+1}$ and the stored memory gradient $g^\ast_k$. If the inner product $\langle g^\ast_k, g_{k+1} \rangle$ is negative, it signifies that the angle between the new gradient and the memory gradient is greater than 90 degrees. In such cases, the new gradient $g_{k+1}$ must be projected onto the nearest gradient $\tilde{g}$ that satisfies $\langle \tilde{g}, g_{k+1} \rangle \geq 0$. This leads to the following optimization problem:
\begin{equation}
\min_{g} \left\| g_{k+1} - g \right\|_2^{2}\quad \text{ s.t. }\quad \langle g,\,g_k^\ast \rangle \geq 0\,,
\end{equation}
with the solution given by:
\begin{equation}
\tilde{g}= g_{k+1} - \frac {\langle g_{k+1},g_k^\ast \rangle}{\left\|g_k^\ast\right\|_2^2}\,g_k^\ast.
\end{equation}

A-GEM places significant emphasis on the new gradient, which may result in issues when the new gradient is too small compared to the gradient of the memory in some specific training epochs \cite{guo2020improved}. To overcome this limitation, a simple modification of A-GEM is developed. This modified approach places emphasis on the gradient of the memory when the gradient of the new data is too small. To do so, when $\langle \tilde{g},\,g_{k+1} \rangle \leq 0$, we condition the new gradient projection to the magnitude of the two gradients. If the magnitude of the new gradient is smaller than the magnitude of the memory gradient, instead of projecting the new gradient, the algorithm projects the gradient of memory to the nearest gradient that satisfies $\langle \tilde{g},\,g_{k+1} \rangle \geq 0$: 
  
\begin{equation}{\begin{cases}
\tilde{g}= g_{k+1} - \frac {\langle g_{k+1},g_k^\ast \rangle}{\left\|g_k^\ast\right\|_2^2}\,g_k^\ast\quad \text{ s.t. }\left\| g_{k+1} \right\| \geq \left\| g_k^\ast \right\|\\\

\\

\tilde{g}= g_k^\ast - \frac {\langle g_{k+1},g_k^\ast \rangle}{\left\|g_{k+1}\right\|_2^2}\,g_{k+1} \quad \text{ s.t. } \left\| g_k^\ast \right\| \geq \left\| g_{k+1} \right\|
\end{cases}}\label{eqProposed}
\end{equation}

The second term in \eqref{eqProposed} ensures that, during certain training epochs where the new gradient is smaller than the memory gradient, greater emphasis is placed on the memory gradient. In both terms of Eq. \eqref{eqProposed}, it can be shown that the inner product between the gradient of the new data and the gradient of the memory points is positive: 

\begin{equation}
    \langle \tilde{g},\,g_k^\ast \rangle \geq 0\
\end{equation}

\section{Multivariate Representative Selection in time series tasks}
The selected memory in the original implementation of the A-GEM optimization algorithm is assumed to represent the best subset of the historical data. The original A-GEM, along with our refined version introduced in the previous section, use a random subset of historical data as memory points, which is not an optimal selection of past memory. Recently, ReSeleCT \cite{hosseinzadeh2024reselect} was proposed, which employs change point detection to identify key locations in the time series data with univariate output. These change points are considered the most informative observations and are selected as memory data. The selection process involves the Narrowest-Over-Threshold (NOT) method, which applies the Gaussian generalized likelihood ratio (GLR) test to detect change points. Employing this change point detection method, the ReSeleCT algorithm iteratively selects representatives from each new dataset (change points) and combines them with the existing memory to optimize the neural network using the A-GEM framework. This process allows the model to learn new information while retaining essential knowledge from previous tasks, leading to improved performance in continual learning scenarios. 
However, a limitation of ReSeleCT is that it is constrained to univariate output time series continual learning scenarios. A generalized approach is required for multivariate cases, where the selection of representative change points must account for the multivariate output environment. 

In EM-ReSeleCT, the focus is on identifying specific changes in the output function pattern for each dimension, as these locations are more informative for time series regression tasks and are particularly vulnerable to catastrophic forgetting. To do so, EM-ReSeleCT applies change point detection methods to find specific landmarks based on each output variable and then applies an error-based filtering of proximal selected points in each output variable to find the best subset and avoid redundant selection. To begin, Let us consider a multivariate setting of $p$ input and $q$ output variables, as in Eq. \ref{Eq1} where $q$ represents the number of outputs to be estimated. Similar to Eq. \ref{AllDataset}, let $D$ denote the dataset of $n$ instances for the pair of input and output vectors, i.e.

\begin{equation}
    D = \left\lbrace \left(\mathbf{x}_1,\,\mathbf{y}_1 \right),\,\dots,\,\left(\mathbf{x}_{n},\,\mathbf{y}_{n} \right) \right\rbrace\,,
\end{equation}
where $\mathbf{x}_j=( x_{1j},\, \dots,\, x_{pj})'$ and $\mathbf{y}_j=(y_{1j}, \,\dots, \,y_{qj})'$ are $p$- and $q$-dimensional vectors of inputs and outputs at time instance $i$, respectively. Additionally, for $i=1,2, \dots, q $, let $\mathcal{Y}_i=\left\lbrace y_{i1},\,y_{i2},\dots,\, y_{in}\right\rbrace$ represents the dataset corresponding to the $i_{th}$ coordinate (variable) of the multivariate time series $\mathbf{y}_1,\, \mathbf{y}_2,\,\dots,\,\mathbf{y}_n$.





\begin{figure}[htbp]
\includegraphics[width=0.5\textwidth]{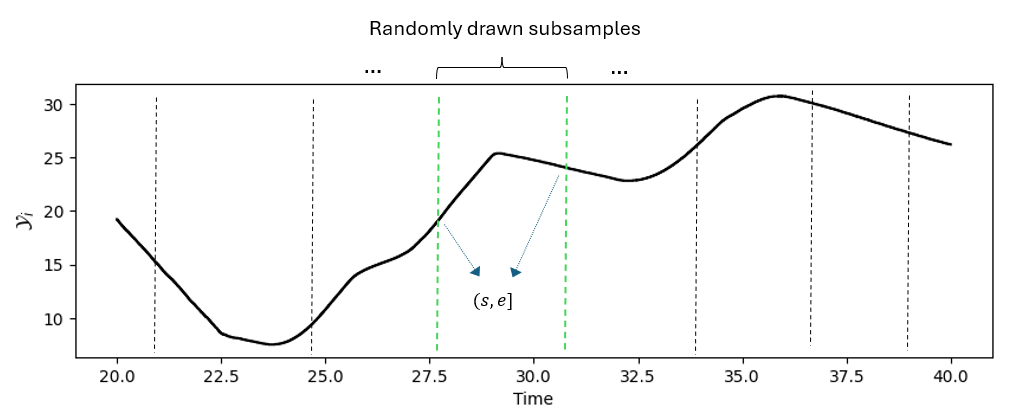} 
\caption{Representative selection on each dimension separately.} 
\label{NOT} 
\end{figure}

The objective is to identify representative points from the dataset $D$ that best capture the original dataset's structure. To achieve this, first, a change point detection method called the Narrowest-Over-Threshold (NOT) \cite{baranowski2019narrowest} is utilized to detect changes in the mean function $\mu_t$ of each output time series $\mathcal{Y}_i$. Let us denote the set of the change points by $\mathcal{Y}^\ast_{i}$ (performing over all dimensions of the output dataset in the historical dataset $\mathcal{Y}$). Therefore, we begin by focusing on selecting a representative subset of the output data for each output variable individually. It is important to note that the choice of the change point detection method depends on both its effectiveness and computational complexity. In this work, we use the NOT algorithm, as it allows for change point selection through a piece-wise linear modeling of the mean of the time series dataset while offering excellent computational efficiency. The NOT algorithm begins by partitioning the dataset $\mathcal{Y}_i$ into $M_i$ randomly selected sub-samples. These sub-samples are determined based on observation indices, as depicted in Figure \ref{NOT}, where the interval \((s,e]\) consists of consecutive observations \(y_{i(s+1)},\,\dots,\,y_{ie}\). Assuming each sub-sample contains at most one change point, the Gaussian Generalized Likelihood Ratio (GLR) test statistic is computed for potential change points within the interval \((s,e]\). If no change point is present, the likelihood function \(L\left ( y_{i(s+1)},\,\dots,\,y_{ie};\,\theta \right )\) is maximized to estimate the intercept and slope parameters \(\eta = (b,\,m)\) of the linear mean function \(\mu_t\) over the range \(t\in (s,e]\). Conversely, if a change point occurs at \(c \in (s,e]\), the mean function \(\mu_t\) exhibits distinct intercepts and slopes in the sub-intervals \((s,c]\) and \((c,e]\), characterized by parameters \(\eta_1\) and \(\eta_2\), respectively. The GLR test statistic, used to compare the likelihood of no change point versus the presence of a change point at \(c\), is formulated as:

\begin{equation}\label{GLR}
\text{\scalebox{0.70}{ $
    \Re _{\left ( s,e  \right   ]}^{c}(\mathcal{Y}_i)= \\ 2\,\log\left [ \frac {\sup\limits_{\eta_1,\eta_{2}}\left \{ L\left ( y_{i(s+1)},...,y_{ic};\eta_{1} \right )L\left ( y_{i(c+1)},\,\dots,\,y_{ie};\eta_2 \right ) \right \}}{\sup\limits_{\eta}L\left ( y_{i(s+1)},\,\dots,\,y_{ie};\eta \right )} \right ] $}}
\end{equation}

Applying the GLR test to each interval \((s,e]\), the NOT method estimates a change point by identifying the value of \(c\) that maximizes equation \eqref{GLR}:

\begin{equation}
    \Re _{\left ( s,e  \right   ]}(\mathcal{Y}_i) = \max_{c\in \left \{ s+2,\dots,e-2 \right \}} \Re _{\left ( s,e  \right   ]}^{c}(\mathcal{Y}_i)
\end{equation}

Thus, within each interval, the change point with the highest likelihood ratio is identified as a potential change point. Once all candidate change points are determined across the \(M_i\) randomly selected intervals, only those with GLR values exceeding a predefined threshold are retained, while the rest are discarded. Finally, among the retained candidates, the change point located within the narrowest interval is selected as the final change point. For further details, refer to \cite{baranowski2019narrowest}.

After gathering all change points (landmarks) from each output dimension, i.e. $\mathcal{Y}^\ast_1, \mathcal{Y}^\ast_2, \dots, \mathcal{Y}^\ast_q$, the algorithm collects their corresponding value from the other output dimensions. This results in $q$ distinct sets, each corresponding to the selected change points based on one variable, along with the corresponding outputs from other dimensions:

\begin{equation}
{\begin{cases} 
    \mathcal{Y}^{**}_{1} = \left( \mathbf{y}^{(1)}_{_1} ,\,\dots,\,\mathbf{y}^{(1)}_{_{m_1}}  \right) \\

    \mathcal{Y}^{**}_{2} = \left( \mathbf{y}^{(2)}_{_1} ,\,\dots,\,\mathbf{y}^{(2)}_{_{m_2}}  \right) \\

    \vdots \\

    \mathcal{Y}^{**}_{q} = \left( \mathbf{y}^{(q)}_{_1} ,\,\dots,\,\mathbf{y}^{(q)}_{_{m_q}}  \right) 
\end{cases}}
\end{equation}

where $\mathcal{Y}^{**}_{i}$ represents the set of representative points selected based on the $i$th output variable in the dataset and $\mathbf{y}^{(i)}_{j} = ( y^{(i)}_{1j}, y^{(i)}_{2j}, \dots, y^{(i)}_{qj} )$ is an observation vector in the dataset where the mean function of $i^{\rm th}$ coordinate (or output variable) corresponding to $j^{\rm th}$ change when only investigating the $i^{\rm th}$ output variable. One potential solution is to identify change points independently for each output, followed by aggregating these points along with their respective inputs and outputs. However, this approach can result in inefficiencies due to the excessive selection of points. This inefficiency arises from the possibility of redundant point selection, as certain points may already have been chosen for a different output variable in a neighboring measurement. This problem becomes particularly pronounced in real-world state estimation applications, where the number of measurements is large, and there is minimal variation between consecutive points.

\begin{figure}[htbp]
\includegraphics[width=0.5\textwidth]{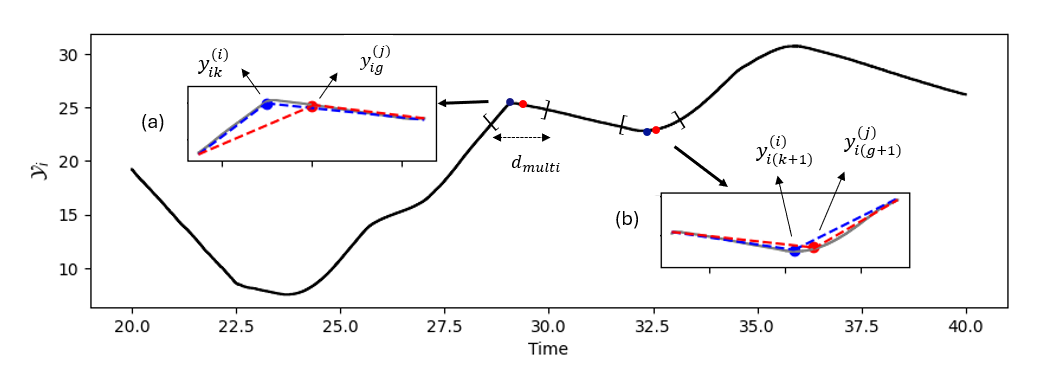} \caption{Multivariate representative selection by filtering based on error fit.} \label{multi} \end{figure}

Figure \ref{multi} illustrates this issue using the $i^{\rm th}$ output time series dataset. In subplot (b), the blue point ($y^{(i)}_{i(k+1)}$) represents the change point detected based on the $i^{\rm th}$ time series output variable, while the red point ($y^{(j)}_{i(g+1)}$) corresponds to a change point detected in another output variable. As shown in the figure, there is no significant difference between selecting either of these points, and selecting both would be redundant. However, this strategy does not apply to subplot (a), where selecting the red point ($y^{(j)}_{ig}$) and discarding the blue one ($y^{(i)}_{ik}$) would result in a significant error.

To address this, we propose a proximal change point selection with minimum error-based filtering. The unique points from each representative set are concatenated. Thus, the concatenated representative set is given by:

\begin{equation}
    \mathcal{Y}^{*c} = \left\lbrace \mathcal{Y}^{**}_{1}, \mathcal{Y}^{**}_{2}, \dots \mathcal{Y}^{**}_{q}\right\rbrace
\end{equation}
 
Let $\mathcal{Y}^{*c}$ represent the set of representative points obtained by concatenating all selected change points for the $j^{\rm th}$ dataset. Now, consider the $i^{\rm th}$ time series output variable in $\mathcal{Y}$ within dataset $D$. Based on equation (16), the set $\mathcal{Y}^*_i$ is identified as the set of change points for this output variable. Suppose the $j^{\rm th}$ point in this set is denoted by $y_{ij}$.

We assume that within the defined distance $d$, only one change point is detected for the $i^{\rm th}$ output variable. This assumption holds for small distances. To evaluate the significance of this change point, we apply the Generalized Likelihood Ratio (GLR) test. Specifically, we perform the GLR test on the segment from $y_{i(j-\frac{d}{2})}$ (the start of the distance) to the selected point $y^{(i)}_{ij}$, and from this point to $y_{i(j+\frac{d}{2})}$ (the end of the distance):

\begin{equation}\label{GLR_yik}
\text{\scalebox{0.67}{ $
    \Re _{\left ( s_1,\,e_1 \right   ]}^{k,\,i}(\mathcal{Y}_i)= \\ 2\,\log\left [ \frac {\sup\limits_{\eta_1,\eta_{2}}\left \{ L\left ( y_{_{i\,(s_1+1)}},...,y_{_{i\,k}};\,\eta_{1} \right )L\left ( y_{_{i\,(k+1)}},\,\dots,\,y_{_{i\,e_1}};\eta_2 \right ) \right \}}{\sup\limits_{\eta}L\left ( y_{_{i\,(s_1+1)}},\,\dots,\,y_{_{i\,e_1}};\eta \right )} \right ] $}}
\end{equation}

where, $s_1=j-\frac{d}{2}$ and $e_1 = j+\frac{d}{2}$, and $y^i_{_{i\,k}}$ is the $k^{\rm th}$ change point detected based on the $i^{\rm th}$ output variable, and the notation $k,\,i$ in eq. (16) represents the GLR test for $k^{\rm th}$ potential change point for output variable $i$. Now, for any change point detected in other output variables, if the corresponding value in the $i^{\rm th}$ variable falls within the corresponding interval, we apply the GLR test on the same interval as follows: 

\begin{equation}\label{GLR_yig}
\text{\scalebox{0.67}{ $
    \Re _{\left ( s_2,\,e_2 \right   ]}^{g,\,j}(\mathcal{Y}_i)= \\ 2\,\log\left [ \frac {\sup\limits_{\eta_1,\eta_{2}}\left \{ L\left ( y_{i(s_2+1)},...,y_{ig};\eta_{1} \right )L\left ( y_{i(g+1)},\,\dots,\,y_{ie_2};\eta_2 \right ) \right \}}{\sup\limits_{\eta}L\left ( y_{i(s_2+1)},\,\dots,\,y_{ie_2};\eta \right )} \right ] $}}
\end{equation}

Note that the notations for Eqs. (\ref{GLR_yik}) and (\ref{GLR_yig}) matches the notations in Figure \ref{multi} in subplot (a). In other words, we applied the GLR test to the interval in subplot (a), one for point $y^{j}_{ig}$ and one for point $y^{i}_{ik}$. We define the error between the two GLR values as:  
\begin{equation}\label{error}
    E = \Re _{\left ( e_1,\,s_1 \right   ]}^{k,\,i}(\mathcal{Y}_i) - \Re _{\left ( e_1,\,s_1 \right   ]}^{g,\,j}(\mathcal{Y}_i)  
\end{equation}

We remove the change point \( y^{j}_{ig} \) from the representative set if the difference between the two GLR values in Eqs. (\ref{GLR_yik}) and (\ref{GLR_yig}) is below a predefined threshold. This is because its effect is negligible, and its neighboring point within the proximity can compensate for its absence. Conversely, if the GLR difference exceeds the threshold, removing \( y^{j}_{ig} \) would significantly impact the representation, so the point is retained in the set.

After selecting the optimal representative set, it is stored in memory and used in conjunction with newly collected data during model training. This process efficiently mitigates the effects of catastrophic forgetting in continual learning models. Algorithm 1 shows the step-by-step procedure for selecting a representative set for memory-based continual learning in multi-output time series scenarios.

\begin{algorithm}
  \DontPrintSemicolon
  \SetAlgoNlRelativeSize{-1}
  {\small
  \KwIn{Historical datasets $D_1,\,\dots,\,D_k$, and the newly collected dataset in new domain $D_{k+1}$, where each \( D_i \) represents the \( i^{\text{th}} \) time series dataset, containing sensor measurements for inputs and corresponding target outputs.}
  \KwOut{Trained model on new data domain, Representative set $D^*$ of historical dataset $D$}

  \BlankLine
  Pre-train the model on the historical dataset $D$. 

  \For{$D_j$ in $\left\{ D_1, D_2, ..., D_k \right\}$ \tcc*{where $D_j$ (eq (3)) consists of sensor measurements $\mathbf{x_j}$ and  $\mathbf{y_j}$ as in eq (1),(2)}}{
    
    \For{$\mathcal{Y}_i$ in $\mathcal{Y} = \left\lbrace \mathcal{Y}_1,\mathcal{Y}_2,\dots,\mathcal{Y}_q \right\rbrace$ \tcc*{where $\mathcal{Y}_i$ is the time series corresponding to the $i$th output dimension}}{
    
      Apply change point detection (NOT method) to the sensor measurements of dataset $\mathcal{Y}_i$. 

      Within a fixed distance $d_{multi}$ for each captured point \( y^{j}_{ig} \) in $i$, calculate the GLR difference based on Eq. \eqref{error}. 

      \If{$E < \text{threshhold}$}{
            remove the captured point \( y^{j}_{ig} \). 
      }
      \Else{
            keep the captured point \( y^{j}_{ig} \).
      }

      $Y^{*M} \leftarrow Y^{*}_i$: Store the captured points from the $i$th dimension along with their associated outputs in other dimensions to memory. 
  
    }

    $D^* \leftarrow D^*_j$: Return and store the final representative set (all input-output pairs as landmarks) for dataset $D_j$ as $D^*_j$.
  }

  Return representative set from all datasets $D^*=\left\{ D^*_1,D^*_2,\dots,D^*_k \right\}$.

  Receive a dataset in the new domain $D_{k+1}$.

  In each training epoch, update the model parameters using the projected gradient: 
  
  \If{$\left\| g_{k+1} \right\| \geq \left\| g_k^\ast \right\|$}{
            $\tilde{g}= g_{k+1} - \frac {\langle g_{k+1},g_k^\ast \rangle}{\left\|g_k^\ast\right\|_2^2}\,g_k^\ast$. 
      }
      
      \Else{
            $\tilde{g}= g_k^\ast - \frac {\langle g_{k+1},g_k^\ast \rangle}{\left\|g_{k+1}\right\|_2^2}\,g_{k+1}$.
      }
    
  Apply steps 3 to 14 when new batch of data arrives.

  Wait for the next time series dataset. 

  \caption{EM-ReSeleCT for Continual Learning}
  \label{algo:sample}
  }
\end{algorithm}

As outlined in the algorithm, after collecting the complete historical dataset from previous domains, a sequence-to-sequence Transformer model (with detailed discussion on its structure in the next section) is initially trained on the first dataset as a pre-trained model. For each historical dataset ($D_j$) in $D$, a multivariate change representative selection method is applied to identify the most representative subset of $D_j$, which consists of selected output points and their corresponding inputs. After gathering all these landmarks into $D^*$ and receiving the new dataset $D_{k+1}$ from the latest domain, the model is retrained using the combined dataset ($D^* \cup D_{k+1}$) with the modified A-GEM optimization algorithm to ensure least forgetting of the previously trained model. Following this continual training on the new dataset, the same representative selection process is applied to the new dataset, with the selected representatives stored for future learning tasks.

\section{Experiments}
\section{Evaluation of the Proposed Method}

\begin{figure}[htbp]
  \centering
  \includegraphics[height=6.5cm,width=0.3\textwidth]{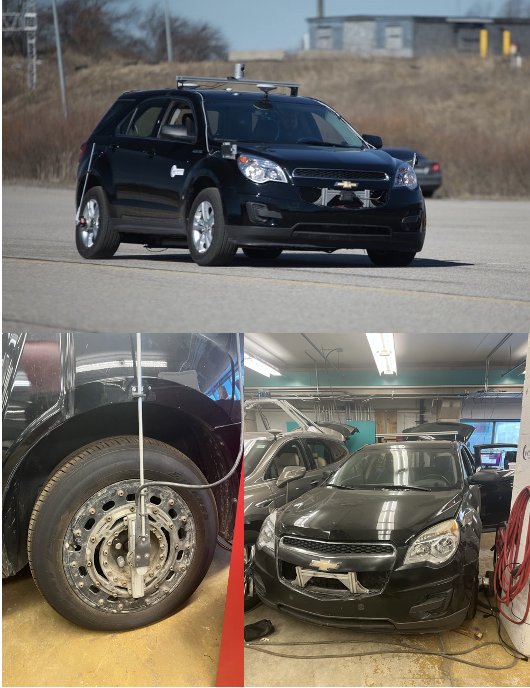}
  \caption{Test vehicle used for experimental analysis.}
  \label{equinox}
\end{figure}

In this section, a comprehensive evaluation of the proposed method, compared to other continual learning approaches, is presented. The objective is to estimate both the longitudinal and lateral velocities of a vehicle using datasets collected through experiments conducted with an electric Equinox vehicle, as illustrated in Figure~\ref{equinox}. Table~\ref{dataset} outlines the specifications of the dataset used in this study. We utilize two sets of training data: an initial, relatively large dataset for training in the initial domain ($D_1$) and a new dataset representing a new domain ($D_2$) that the model must learn. Additionally, one test maneuver from $D_1$ is included to assess the impact of catastrophic forgetting when introducing the new dataset ($D_2$) into the model. As shown in Table~\ref{dataset}, the initial training dataset comprises 10 maneuvers, while the new dataset consists of one maneuver under different road conditions. Measurements were recorded at a frequency of 100 HZ, capturing various sensor data such as yaw rate, wheel speeds, etc.

This section is divided into three subsections. The first subsection details the procedure for the initial training of the neural network model, including hyperparameter settings, input feature selection, and training on the initial dataset. In Subsection~B, the continual learning scenario is described, where various methods for continually learning the new dataset ($D_2$) in a new domain, while retaining knowledge from the initial domain, are implemented and evaluated. Subsection~C presents a comparison of the performance of the proposed and A-GEM algorithm on memory loss minimization.
  
\begin{figure}[]
  \centering
  \includegraphics[height=5.6cm,width=0.47\textwidth]{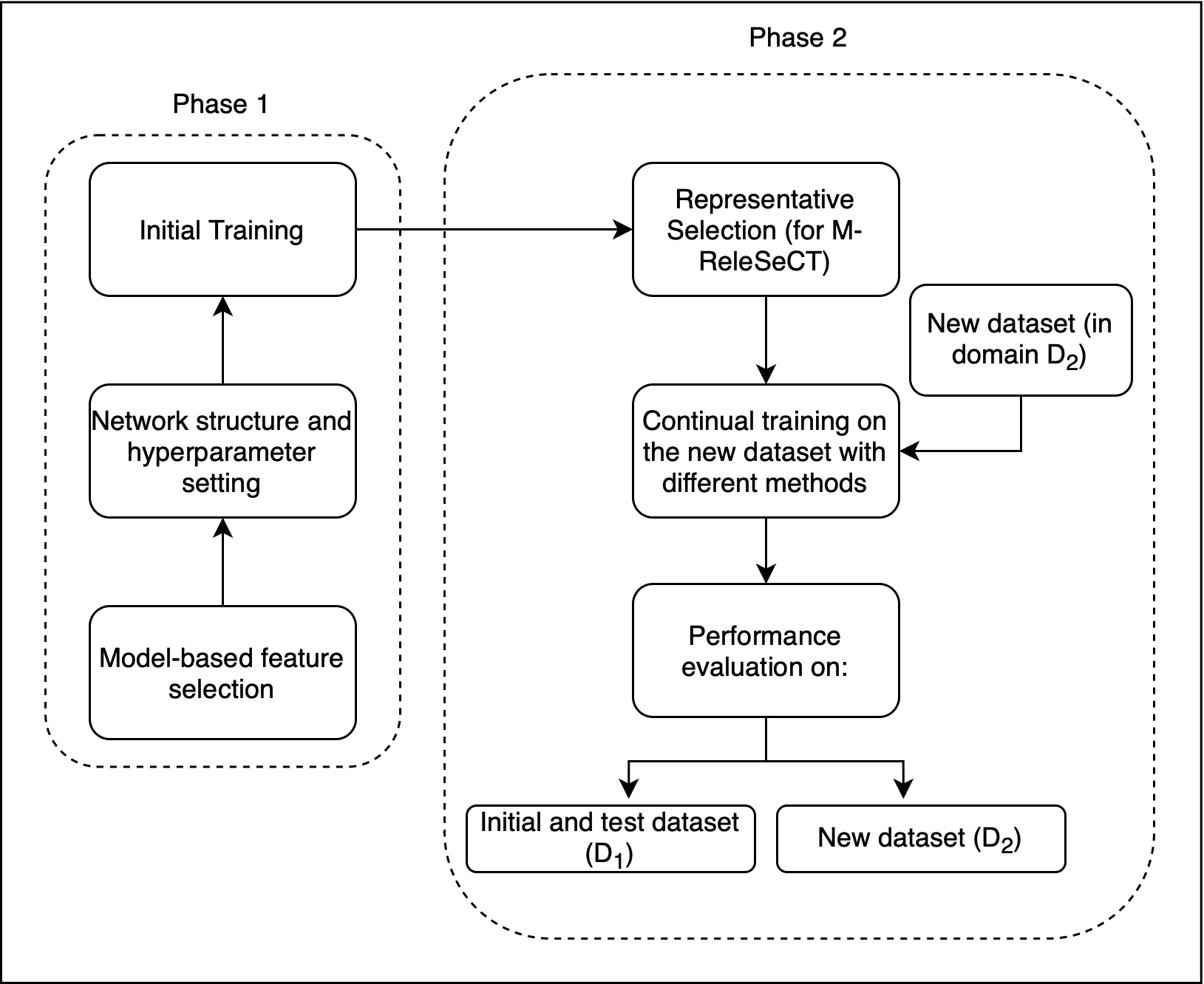}
  \caption{The flowchart of evaluation process of different methods in continual learning scenarios.}
  \label{procedure}
\end{figure} 

\begin{table}[h]
\centering
\caption{Dataset used for evaluation of different methods to estimate vehicle velocity in a continual learning scenario.}
\begin{tabular}{|c|c|c|c|}
\hline
Data type          & Weather condition                     & \# of maneuvers & \# of data \\ \hline
Training:$D_1$ & Sunny(high $\mu$)                 & 10              & 12000            \\  \hline
New Training:$D_2$ & Rainy(low $\mu$)                 & 1              & 1200            \\  \hline

Testing:$D_1$ & Sunny(high $\mu$)                 & 1              & 1200            \\  \hline
\end{tabular}
\label{dataset}
\end{table}

\subsection{Phase 1: Initial Training} 

In the initial step of the evaluation, we train the model parameters using the initial dataset. This step is common across all continual learning techniques, and the same trained model is used in the subsequent sections when new maneuvers are added. We focus on thoroughly training the model with the initial dataset to analyze the impact of forgetting on continual learning. As shown in Table~\ref{dataset}, the initial dataset comprises 10 maneuvers. The objective is to estimate both longitudinal and lateral velocities of the vehicle using input measurements and predicted outputs from previous steps (autoregressive structure) over the period of maneuver time. 

\subsubsection{Feature Selection}
To select the relevant inputs for estimating longitudinal ($V_x$) and lateral ($V_y$) velocities, in addition to the last predicted states of both $V_x$ and $V_y$ using the neural network model, we selected eight different signals: four wheel speeds, yaw rate, steering angle, and longitudinal and lateral accelerations, based on the vehicle model's dynamic relation, as described in \cite{imsland2006vehicle}. 

\subsubsection{Neural Network Model and Hyperparameter Set}

This work adapts the original Transformer model to time series state estimation with the encoder-decoder mechanism. In the Encoder block, the multidimensional input data as selected in the previous subsection is fed into the model using a time-delay embedding (TDE) format based on a predefined window size. Let \( p =8\) represent the dimensionality of each input time step and \( d \) the time delay for the entire sequence. We considered a time delay of 0.5 seconds (50 measurements) for this experimental analysis. The input to the Decoder block consists of the last predicted output states of the model, allowing it to use previously predicted states alongside the output of the Encoder layer in the attention head. The decoder inputs are in a $q$-dimensional space, $q=2$ as we aim to estimate $V_x$ and $V_y$, with a time delay, which is set to match the time delay, $d$, of the encoder inputs. To align with the input space, the decoder inputs are first embedded into the same dimension as the encoder inputs using a linear embedding layer, mapping the $q$-dimensional outputs to $p$ dimensions over all time sequences (time delays). Finally, the output of this block is processed through a feedforward layer with ReLU activation to introduce nonlinearities, followed by a linear layer that maps the preceding outputs to the final output layer, which has a dimension corresponding to the number of predicted outputs ($q=2$). The specifications of the hyperparameters for the Transformer model are outlined in Table \ref{Hyperparams}. 

\begin{table}[h]
\centering 
\caption{Hyperparameter setting of the Transformer model for estimation of $V_x$ and $V_y$.}
\begin{tabular}{|c|c|c|}
\hline
Label & Description                           & Value \\ \hline
$l_e$            & Encoder hidden layers               & 2     \\ \hline
$l_d$            & Decoder hidden layers                & 2   \\ \hline 
$A_t$            & Attention heads                & 2   \\ \hline 
$h$            & Feedforward hidden units                & 500   \\ \hline 
$d$            & Window size (time delay embedding)               & 50    \\ \hline
$m_i$       & Decay rate for the second moment estimate  & 0.9   \\ \hline
$v_i$          & Decay rate for the first moment estimate & 0.95  \\ \hline
$\alpha$       & Learning rate for initial training    & 0.01  \\ \hline
$\alpha^{'}$ & Learning rate for continual learning  & 0.001      \\ \hline
\end{tabular}
\label{Hyperparams}
\end{table}

\subsubsection{Training and estimation results in initial training}

\begin{figure}[]
  \centering
  \includegraphics[height=6cm,width=0.5\textwidth]{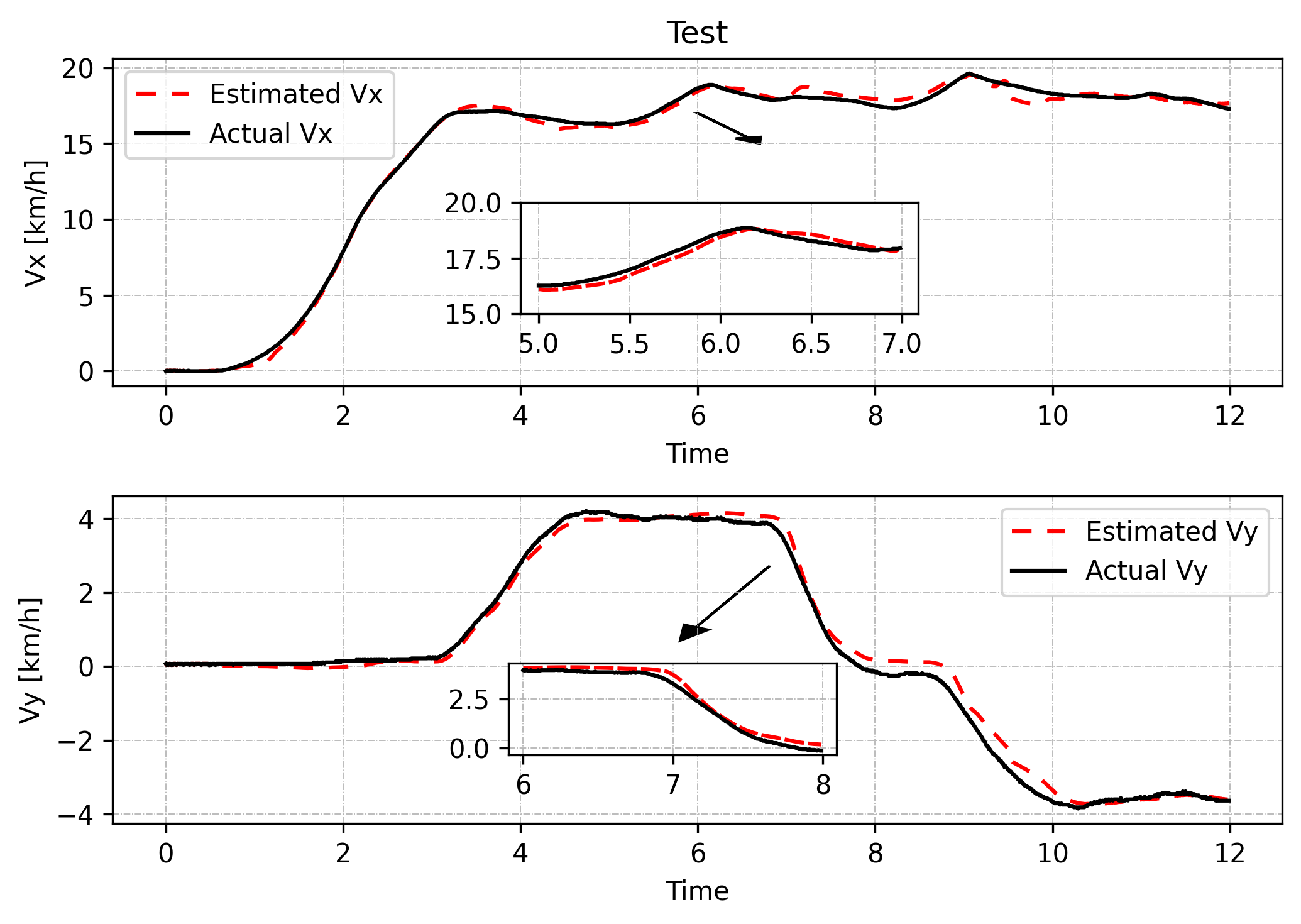}
  \caption{Estimation of $V_x$ and $V_y$ on the test data (domain $D_1$), after the initial training.}
  \label{test_before}
\end{figure}

During initial training, the model is stopped once the MSE loss falls below a predefined threshold. As previously mentioned, we use an autoregressive Transformer model to estimate each successive step of the maneuver. In this setup, the model feeds its prior output estimation into the input decoder rather than the actual output values, and estimation continues until the maneuver completes (typically 12 seconds in our experiments). For the test data, as noted, experiments were conducted within the same domain as the initial dataset to later assess the effect of catastrophic forgetting upon domain expansion. Table~\ref{initial_before} presents the training time and estimation results on both the initial and test datasets, while Figure~\ref{test_before} illustrates the model’s estimation results on the test data within domain $D_1$ following initial training.  

\begin{table}[]
\caption{Estimation results of $V_x$ and $V_y$ on initial and test dataset after initial training ($D_1$).}
\begin{tabular}{|c|c|c|cc|}
\hline
Dataset                          & Maneuver                                   & Training time (s)       & \multicolumn{2}{c|}{MAE (km/h)}    \\ \hline
\multirow{2}{*}{Initial } & \multirow{2}{*}{10 manevuers in $D_1$}      & \multirow{2}{*}{247} & \multicolumn{1}{c|}{$V_x$} & 0.31 \\ \cline{4-5} 
                                 &                                            &                      & \multicolumn{1}{c|}{$V_y$} & 0.12 \\ \hline
\multirow{2}{*}{Test}            & \multirow{2}{*}{drvining at high steering} & \multirow{2}{*}{--}  & \multicolumn{1}{c|}{$V_x$} & 0.37 \\ \cline{4-5} 
                                 &                                            &                      & \multicolumn{1}{c|}{$V_y$} & 0.17 \\ \hline
\end{tabular} \label{initial_before}
\end{table}

As shown in Table~\ref{initial_before} and Figure~\ref{test_before}, the model has an acceptable performance in terms of estimation error, as the maneuvers were performed in the same data domain as the initial dataset, there is no need to adapt the model to the test data.

\subsection{Phase 2: Continual Learning} 

In this section, we introduce a new dataset, collected in a different domain ($D_2$) compared to the initial and test datasets ($D_1$). Figure~\ref{new_before} presents the model’s estimation results on the new dataset before incorporating the new dataset into the model (prior to continual training): 

\begin{figure}[]
  \centering
  \includegraphics[height=6cm,width=0.5\textwidth]{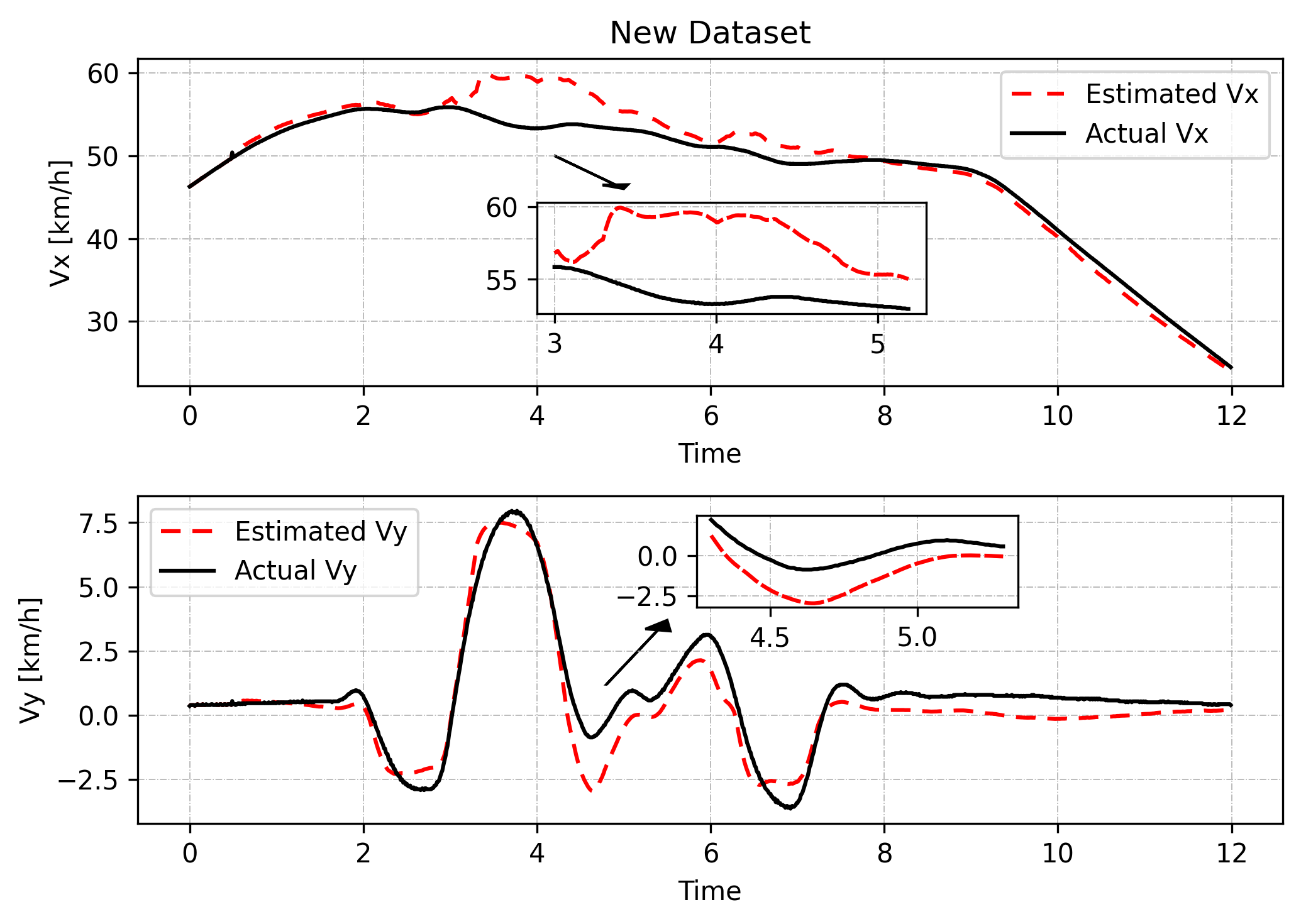}
  \caption{Estimation of $V_x$ and $V_y$ on the new data (domain $D_2$), prior to continual learning.}
  \label{new_before}
\end{figure} 

As shown in the figure, the model performs poorly on the new dataset. This result is expected, as the maneuver was conducted under significantly different road conditions (heavy rain), requiring the model to adapt to the new dataset. To incorporate the new dataset information into the model, we evaluate the proposed EM-ReSeleCT alongside other methods for comparison, as follows: 

\textbf{Batch:} Retrains all neural network parameters using both previously collected and newly acquired data, serving as a baseline or upper bound. This approach, referred to as Batch or Joint training mode.   

\textbf{None:} Trains the model solely on the new maneuver, disregarding previously trained parameters, which represents a lower bound.  

\textbf{A-GEM \cite{chaudhry2018efficient}:} Implements the original A-GEM method, in which memory is selected by randomly sampling from the previous task. 

\textbf{SI \cite{zenke2017continual}:} Adds a regularization term to the cost function to preserve significant weights of the neural network, based on the sensitivity of each parameter.

We consider a specified number of points as memory points for A-GEM ($m_1 = 150$), using the same quantity for EM-ReSeleCT. However, the key difference in EM-ReSeleCT is the selection of a specific representative set rather than random selection. This multivariate analysis within EM-ReSeleCT's structure significantly reduces the number of memory points required. In our approach, after processing historical data with EM-ReSeleCT, the total number of memory points is reduced to $m_1 = 97$. 

Table~\ref{new_after} presents the training time and Mean Absolute Error (MAE) of each continual learning strategy on the new dataset after introducing the domain into the model. As shown, all strategies demonstrate strong performance in estimation error for the new maneuver. In terms of training time, the Batch mode requires substantial time due to its use of all historical data combined with the new data during training. Interestingly, despite the additional computations for memory selection in EM-ReSeleCT and A-GEM, they exhibit comparable training times. This efficiency is attributed to the significantly reduced final memory set in EM-ReSeleCT compared to A-GEM or random selection strategies. SI and None modes have the same training time, as neither retains memory points in the continual learning phase.

\begin{table}[]\centering
\caption{MAE and training time of $V_x$ and $V_y$ estimation on the new maneuver ($D_2$) following model training.}
\begin{tabular}{|c|c|c|cc|}
\hline
\multirow{2}{*}{Maneuver} & \multirow{2}{*}{Method} & \multirow{2}{*}{Training time} & \multicolumn{2}{c|}{MAE}                                \\ \cline{4-5} 
                          &                         &                                & \multicolumn{1}{l|}{$V_x$} & \multicolumn{1}{l|}{$V_y$} \\ \hline
\multirow{5}{*}{$D_2$}    & Batch                 & 65.2                            & \multicolumn{1}{c|}{0.81}  & 0.17                       \\ \cline{2-5} 
                          & None                    & 7.1                            & \multicolumn{1}{c|}{0.70}  & 0.11                       \\ \cline{2-5} 
                          & SI                      & 8.0                           & \multicolumn{1}{c|}{0.74}  & 0.14                       \\ \cline{2-5} 
                          & A-GEM                   & 10.5                          & \multicolumn{1}{c|}{0.84}  & 0.14                       \\ \cline{2-5} 
                          & EM-ReSeleCT             & 10.9                           & \multicolumn{1}{c|}{0.71}  & 0.15                       \\ \hline
\end{tabular}\label{new_after}
\end{table}

\begin{figure}[]
  \centering
  \includegraphics[height=6cm,width=0.5\textwidth]{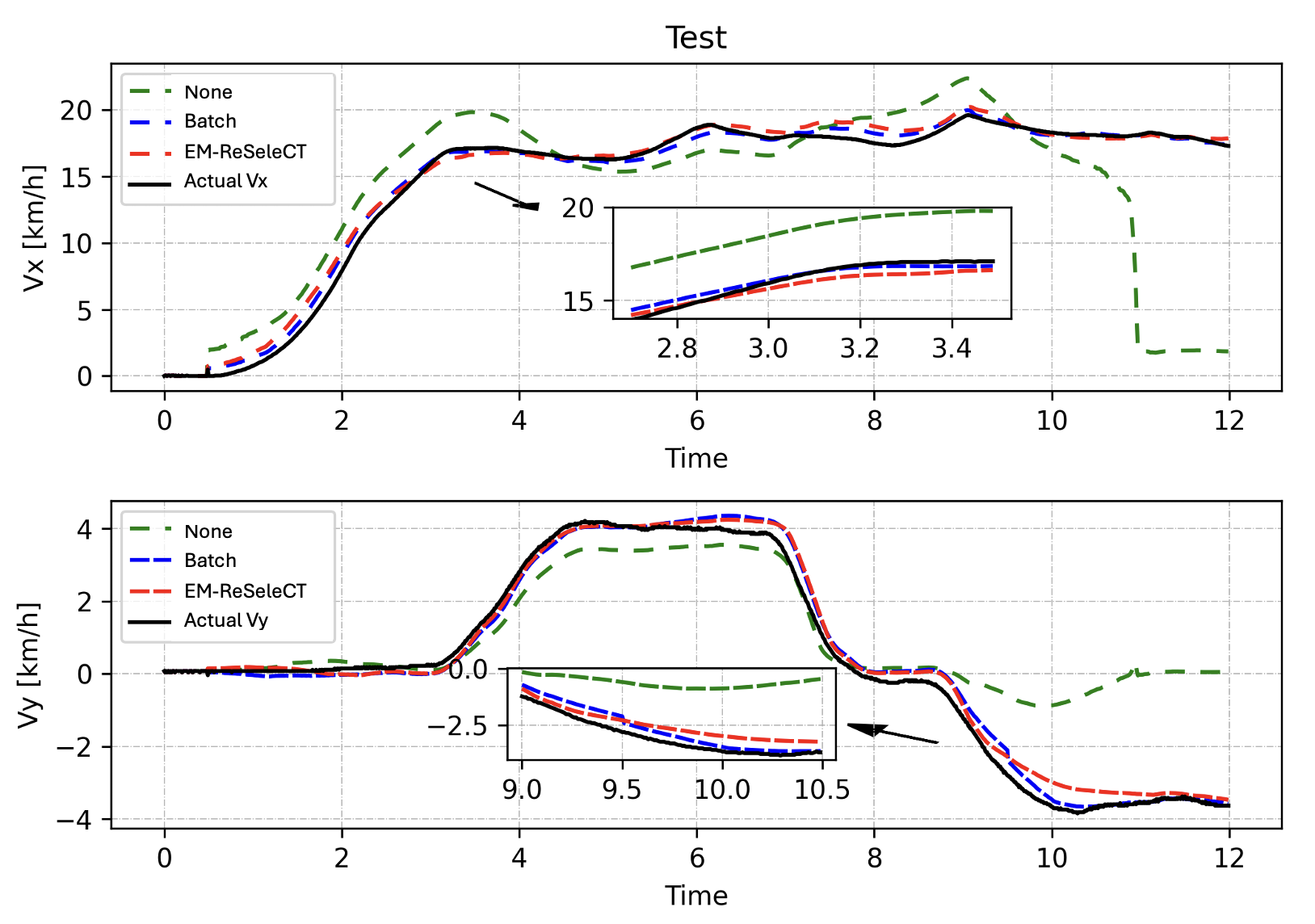}
  \caption{$V_x$ and $V_y$ estimations on the test maneuvers (domain $D_1$), after incorporating the new dataset (in $D_2$).}
  \label{test_aftercompare1}
\end{figure}

\begin{figure}[]
  \centering
  \includegraphics[height=6cm,width=0.5\textwidth]{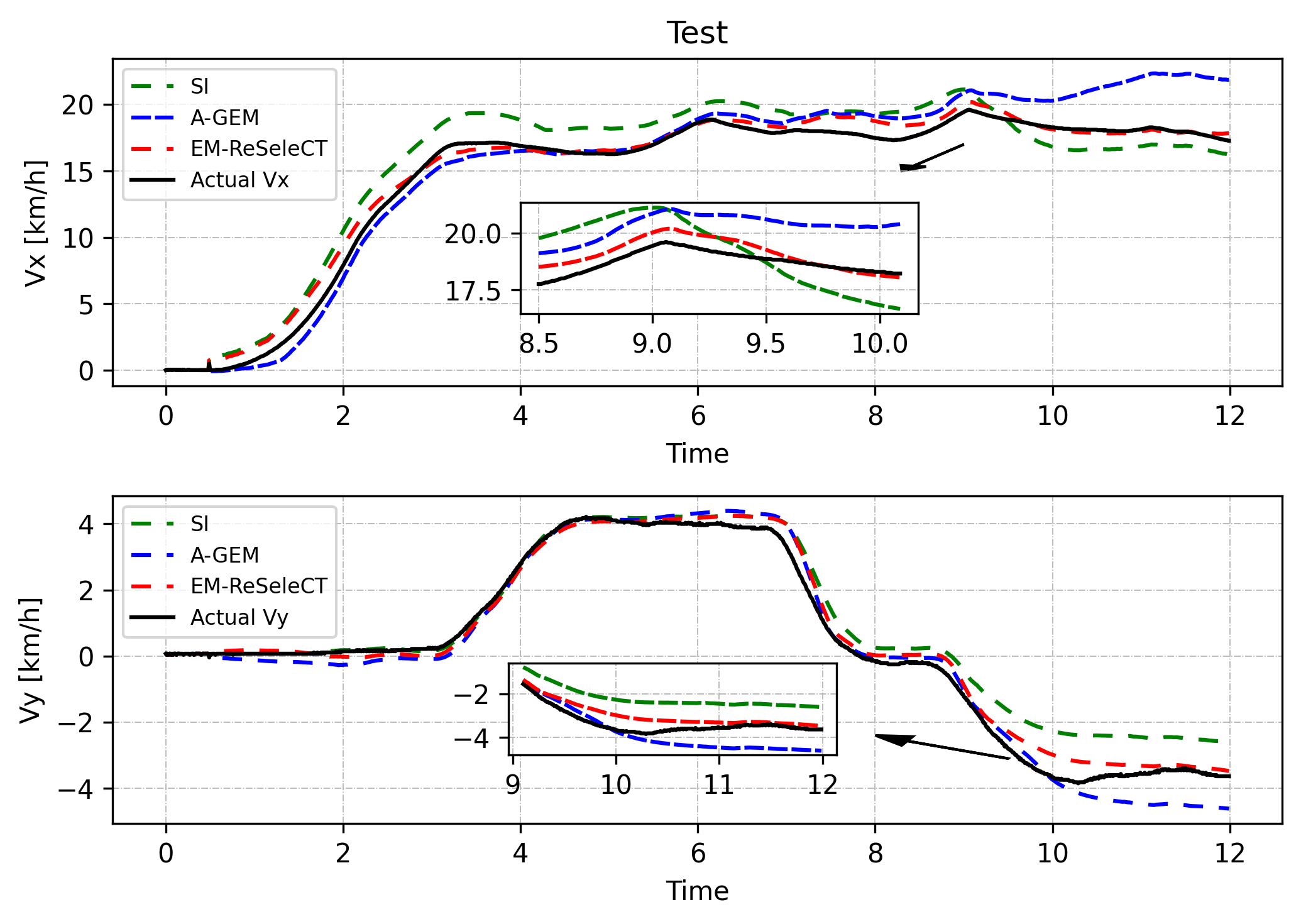}
  \caption{Evaluation of the proposed method against other continual learning (CL) strategies for estimating \( V_x \) and \( V_y \) on test data after incorporating the new dataset (\( D_2 \)) into the model.}
  \label{test_aftercompare2}
\end{figure}

A critical aspect of continual learning techniques is their capacity to preserve previously learned knowledge across domains encountered so far. Table~\ref{test_new} and Figures~\ref{test_aftercompare1} and \ref{test_aftercompare2} display the estimation results on the test data (in $D_1$) using different continual learning strategies, following the introduction of a new domain (in $D_2$) to the model. As shown, the proposed method (EM-ReSeleCT) achieves an estimation error comparable to that of the Batch mode, which serves as our baseline. EM-ReSeleCT significantly outperforms other popular CL strategies, such as A-GEM and SI. This superior performance over A-GEM is attributed to EM-ReSeleCT’s capability to select informative memory points rather than relying on random selection, coupled with optimization modifications that enable positive backward transfer, enhancing learning on previously encountered data. SI, on the other hand, shows poor performance, as weight regularization methods have consistently demonstrated limited accuracy in these regression problems.

\begin{table}[] 
\centering 
\caption{MAE of $V_x$ and $V_y$ estimation on training and testing maneuvers ($D_1$) after introducing the new maneuver ($D_2$).}
\begin{tabular}{|c|cc|cc|}
\hline
\multirow{2}{*}{Method} & \multicolumn{2}{c|}{Initial Dataset} & \multicolumn{2}{c|}{Test}          \\ \cline{2-5} 
                        & \multicolumn{1}{c|}{$V_x$}  & $V_y$  & \multicolumn{1}{c|}{$V_x$} & $V_y$ \\ \hline
Batch                 & \multicolumn{1}{c|}{0.33}    & 0.17    & \multicolumn{1}{c|}{0.36}  & 0.15  \\ \hline
None                    & \multicolumn{1}{c|}{4.21}    & 1.69    & \multicolumn{1}{c|}{3.09}  & 1.00  \\ \hline
SI                      & \multicolumn{1}{c|}{1.4}   & 0.95   & \multicolumn{1}{c|}{1.56}  & 0.44  \\ \hline
A-GEM                   & \multicolumn{1}{c|}{0.79}  & 0.35   & \multicolumn{1}{c|}{1.32}  & 0.31  \\ \hline
EM-ReSeleCT             & \multicolumn{1}{c|}{0.47}   & 0.25  & \multicolumn{1}{c|}{0.41}  & 0.23  \\ \hline
\end{tabular} 
\label{test_new}
\end{table}  

Figure~\ref{test_aftercompare1} presents the estimation results of EM-ReSeleCT in comparison to the Batch and None modes. As illustrated, for both longitudinal ($V_x$) and lateral ($V_y$) velocities, the proposed method demonstrates promising performance, closely aligning with the Batch mode but with significantly lower computational time. In specific regions, EM-ReSeleCT effectively preserves information, whereas in the None mode—where prior information is disregarded—the model performs poorly.

Figure~\ref{test_aftercompare2} further compares the estimation results of EM-ReSeleCT with A-GEM and SI. As shown, EM-ReSeleCT exhibits superior performance, while both A-GEM and SI struggle to retain historical information (in $D_1$). Over time, this results in divergence, causing estimation accuracy for $V_x$ and $V_y$ to deteriorate substantially. 

\subsection{EM-ReSeleCT v.s. A-GEM: Performance on memory loss}

\begin{figure}[]
  \centering
  \includegraphics[height=4.8cm,width=0.42\textwidth]{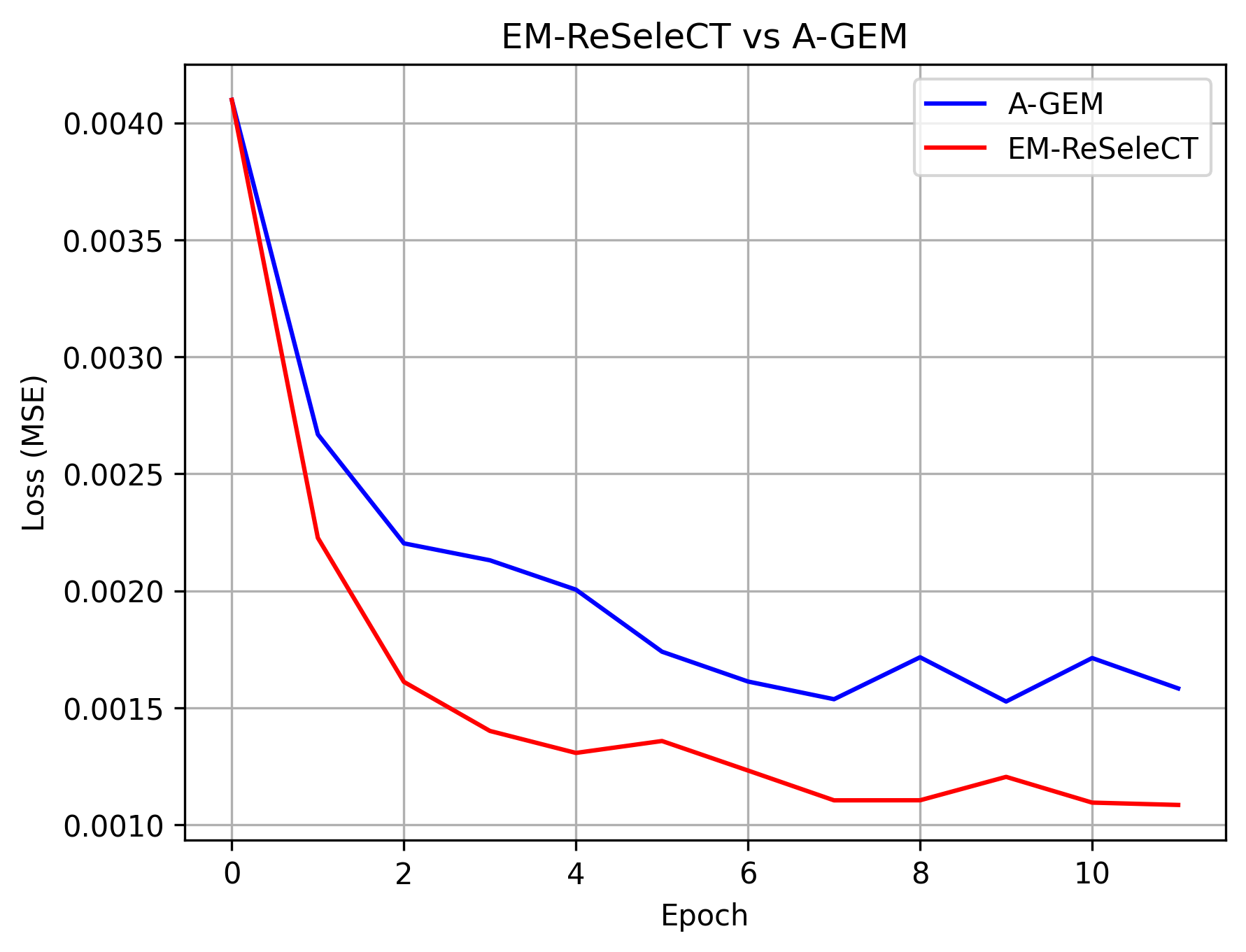}
  \caption{Tracking the loss error over each training epoch in continual learning with EM-ReSeleCT and A-GEM.}
  \label{loss_comparison}
\end{figure}

As previously mentioned, EM-ReSeleCT enhances A-GEM’s optimization algorithm to better preserve historical data, as outlined in Eq. \eqref{eqProposed}. Figure~\ref{loss_comparison} illustrates the loss on memory points across training epochs in a continual learning scenario. Our algorithm not only prevents an increase in memory point loss but also reduces it over certain epochs, demonstrating improved retention of historical information. In addition to the modified optimization, EM-ReSeleCT selects key information by identifying informative data points rather than using A-GEM’s random selection, which further mitigates catastrophic forgetting across all historical data.  

\subsection{Uncertainty Quantification with Conformal Prediction}

Validating the performance of models with uncertainty analysis is crucial for assessing their reliability. Conformal prediction (CP) is a statistical framework that provides calibrated prediction intervals for model outputs, ensuring a specified level of confidence. In this work, CP is employed to quantify the uncertainty of the proposed model on both the old test dataset and the new dataset. Specifically, CP is used to evaluate the uncertainty and coverage rate of the proposed model under different numbers of memory points. It is important to note that we use split conformal prediction (Vainla CP) as described in \cite{vovk2005algorithmic}, which assumes a fixed set of non-conformity scores and a non-exchangeable pair of inputs and outputs. We argue that this assumption holds in our context, as the calibration set, after introducing the new maneuver, incorporates both the old and new datasets, and there is no distribution shift between the calibration and test set. 

Given a prediction model \(f(x)\), conformal prediction constructs prediction intervals based on the residuals from a calibration set. Let \(\mathcal{D}_{\text{train}}\), \(\mathcal{D}_{\text{calib}}\), and \(\mathcal{D}_{\text{test}}\) represent the training, calibration, and test datasets, respectively. For each calibration point \((x_i, y_i) \in \mathcal{D}_{\text{calib}}\), the residual is calculated as:
\begin{equation}
    r_i = y_i - \widehat{f}(x_i), \quad i = 1, \dots, n.
\end{equation}

The quantile of the residuals at level \(1 - \alpha\) is computed as:
\begin{equation}\label{quantile}
    q_{1-\alpha} = \text{Quantile}_{1-\alpha} \{r_1, r_2, \dots, r_n\}.
\end{equation}

For a new test input \(x_{\text{test}}\), the prediction interval is then given by:
\begin{equation}
    \text{PI}(x_{\text{test}}) = \left[\widehat{f}(x_{\text{test}}) - q_{1-\alpha}, \widehat{f}(x_{\text{test}}) + q_{1-\alpha}\right],
\end{equation}

where $\widehat{f}(x_{\text{test}})$ is the estimated output for input $x_{test}$, and $q_{1-\alpha}$ is the quantile calculated in Eq~\ref{quantile} This interval is guaranteed to cover the true target value \(y_{\text{test}}\) with probability at least \(1 - \alpha\), under the assumption that the calibration set is exchangeable with the test data. In this study, we applied conformal prediction to estimate the prediction intervals for the outputs \(V_x\) and \(V_y\). Separate quantiles were computed for each output dimension, providing tailored uncertainty bounds for multivariate predictions. 

To ensure reliable uncertainty quantification, we recalibrate the conformal prediction model after introducing the new dataset. The calibration data includes samples from both the old and new datasets to ensure comprehensive coverage. After determining the quantile $q_\alpha$ for each model, we apply the uncertainty quantification to both the test data from the old data space and the new data space. 

\begin{figure*}[htbp]
    \centering
    \includegraphics[width=0.8\textwidth]{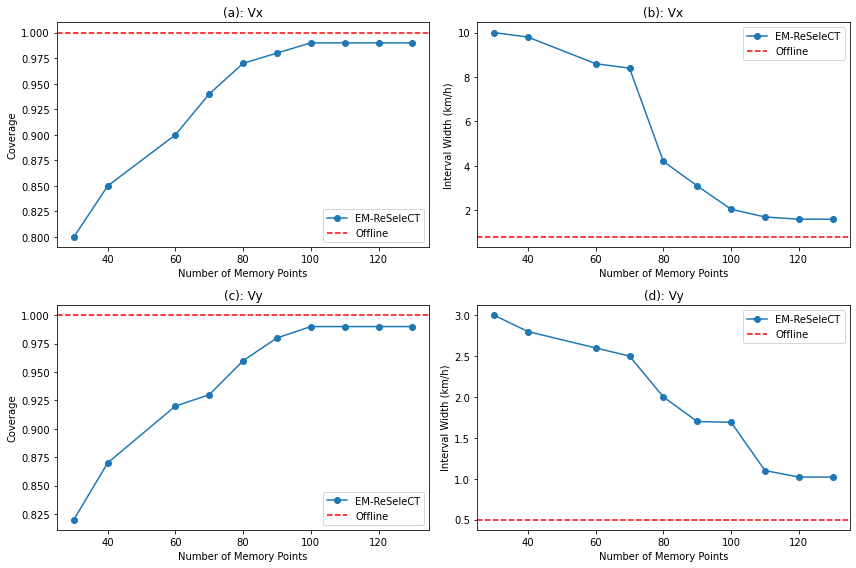}
    \caption{Uncertainty analysis of different models by number of selected memory points, on the old (test) dataset. a and b: coverage rate and interval width vs the number of memory points on $Vx$, respectively. c and d: coverage rate and interval width vs the number of memory points on $Vy$, respectively.}
    \label{cp}
\end{figure*}

Figure~\ref{cp} (a) and (b), show the coverage rate and interval width for the estimation of $V_x$ with different numbers of memory points selected for the proposed model on the test dataset from the old data space. 

Figure~\ref{cp} (c) and (d), illustrate the coverage rate and interval width for the estimation of $V_y$ with different numbers of memory points selected for the proposed model.
 
As illustrated in the figures, the interval size and uncertainty level increase significantly as the number of memory points decreases. Furthermore, the figures demonstrate that after increasing the number of memory points beyond a certain threshold, the interval size remains unchanged. This suggests that storing additional memory points beyond this threshold does not contribute to further efficiency gains. The reason is that the memory points selected by the automatic approach discussed in the previous section act as representatives of historical data. Over-selecting leads to retaining redundant, non-informative samples that neither enhance information preservation nor effectively mitigate catastrophic forgetting. 
 
\section{Conclusions and Future Work}

In this paper, we introduce EM-ReSeleCT, an efficient multivariate representative selection and optimization strategy for continual learning in time series tasks. Experimental analysis on multivariate output estimation using Transformer models, applied to the electric Equinox vehicle, demonstrates the superior performance of EM-ReSeleCT in both training time and estimation accuracy compared to other state-of-the-art continual learning algorithms. Specifically, EM-ReSeleCT achieves estimation errors close to the Batch mode while significantly reducing computational time. The method effectively minimizes the number of memory points by identifying instances where two selected points across outputs represent the same information but are in close proximity.

EM-ReSeleCT is versatile and applicable to other time series domains, including electric vehicle battery management and financial market forecasting. Additionally, it holds potential for adaptation to other data types, enabling performance gains in classification tasks, which is a focus of future research by the authors. 

\section*{Acknowledgments}
The authors would like to acknowledge the financial support of the Natural Sciences and Engineering Research Council of Canada in this work.
\bibliographystyle{IEEEtran}
\bibliography{M-ReSeleCT.bib}

\vfill

\end{document}